\documentclass[10pt,twocolumn,letterpaper]{article}

\usepackage[pagenumbers]{iccv} % To force page numbers, e.g. for an arXiv version

\usepackage[pagebackref,breaklinks,colorlinks,allcolors=blue]{hyperref}
\usepackage{multirow}
\usepackage{xcolor}
\usepackage{amsfonts}
\usepackage{pifont}
\usepackage{enumitem}
\usepackage{algorithm}   
\usepackage{algorithmic} 
\usepackage{adjustbox, bbm}
\def\paperID{*****} % *** Enter the Paper ID here
\def\confName{ICCV}
\def\confYear{2025}

\title{Index-Aligned Query Distillation for Transformer-based Incremental Object Detection}

\author{Mingxiao Ma, Shunyao Zhu, Guoliang Kang\thanks{Corresponding author}\\
Beihang University \\
\tt \small mamingxiao@buaa.edu.cn, shunyao.zhu@outlook.com, guoliang\_kang@buaa.edu.cn
}

\begin{document}
\maketitle
\begin{abstract}
Incremental object detection (IOD) aims to continuously expand the capability of a model to detect novel categories while preserving its performance on previously learned ones. When adopting a transformer-based detection model (\emph{e.g.,} DETR) to perform IOD, catastrophic knowledge forgetting may inevitably occur, which means the detection performance on previously learned categories may severely degenerate. Previous typical methods mainly rely on knowledge distillation (KD) to mitigate the catastrophic knowledge forgetting of transformer-based detection models in IOD. Specifically, they utilize Hungarian Matching to build a correspondence between the queries of the last-phase and current-phase detection models and align the classifier and regressor outputs between matched queries to avoid knowledge forgetting. However, we observe that in IOD task, Hungarian Matching is not a good choice. With Hungarian Matching, the query of the current-phase model may match different queries of the last-phase model at different iterations during KD. As a result, the knowledge encoded in each query may be reshaped towards new categories, leading to the forgetting of previously encoded knowledge of old categories. Based on our observations, we propose a new distillation approach named Index-Aligned Query Distillation (IAQD) for transformer-based IOD. Beyond using Hungarian Matching, IAQD establishes a correspondence between queries of the previous and current phase models that have the same index. Moreover, we perform index-aligned distillation only on partial queries which are critical for the detection of previous categories. In this way, IAQD largely preserves the previous semantic and spatial encoding capabilities without interfering with the learning of new categories. Extensive experiments on representative benchmarks under different IOD protocols demonstrate that IAQD effectively mitigates knowledge forgetting, achieving new state-of-the-art performance.
\end{abstract}    
\section{Introduction}
\label{sec:intro}
Object detection has made remarkable progress~\cite{ren2015faster,jiang2022review,carion2020end}. Among them, transformer-based detectors, \emph{e.g.,} DETR~\cite{carion2020end}, Deformable DETR~\cite{zhu2020deformable}, and other DETR variants~\cite{dai2021up, li2022dn, liu2022dab, meng2021conditional,zhang2022dino, zhao2024detrs}, become representative detection models in practice due to their superior detection performance and end-to-end advantage.  
These detection methods are designed for scenarios where annotations of full categories are accessible. However, in the real world, new categories usually constantly emerge~\cite{caccia2021new, joseph2021towards}. 
The conventional transformer-based detection models, trained with categories emerging at the previous phase, struggle with catastrophic forgetting after being trained with new categories emerging at the current phase~\cite{liu2023continual, kang2023alleviating}, which means the detection performance on old categories may dramatically decrease. 
Thus, incremental object detection (IOD) has attracted great focus from researchers and practitioners. IOD models are designed to learn new categories sequentially while retaining knowledge of previously learned ones~\cite{shmelkov2017incremental}. 

\begin{figure*}[h]
    \centering
    \includegraphics[width=1.0\linewidth]{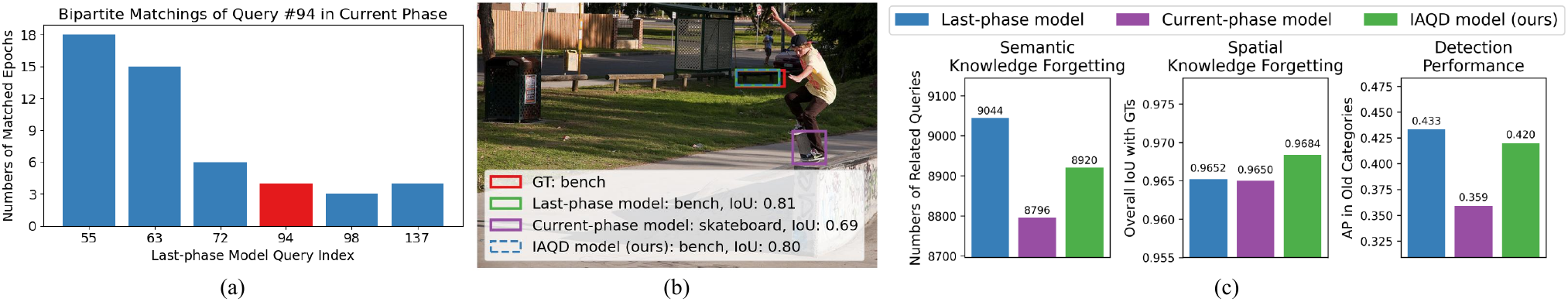}
    \caption{We utilize the last-phase and current-phase models in CL-DETR, as well as our IAQD model. Specifically, 70 categories are learned in the last phase and 10 in the current phase. (a) shows the numbers of matchings between query \#94 in the current-phase model and queries with all indexes in the last-phase model during the 50-epoch training. 
    (b) shows the detections of query \#94 in different models and the corresponding GT of the last-phase model. 
    (c) In the left figure, the ``related queries'' refer to the queries bipartite-matched with the foregrounds during training. The numbers, obtained by summing the numbers of related queries for old categories, reflect the quality of the semantic information encoded in the queries. In the middle figure, the overall IoU is the average of IoUs computed for each old category: for each category, the IoU between the overall regions covered by related query-generated bounding boxes and their corresponding GT regions. This can reveal the global spatial information encoded in the queries. The right figure shows the APs of different models for detecting old categories.}
    \label{query}
\end{figure*}
Recent approaches in IOD typically address the issue of catastrophic forgetting through knowledge distillation (KD)~\cite{hinton2015distilling, douillard2020podnet, hu2021distilling, hou2019learning, li2017learning, zhao2020maintaining, zhang2025dynamic, liu2023continual, wang2025psedet, wang2025gcd, bai2025class, bai2024revisiting}. In a transformer-based detection model, queries are assumed to encode the semantic and spatial information of objects to be detected.
Thus, previous transformer-based IOD methods choose to distill the knowledge encoded by queries of the previous-phase model to the queries of the current-phase model.
To enable the query distillation, an important step is to build the query correspondence between the previous-phase and current-phase models.
Previous work typically uses Hungarian Matching to establish the correspondence~\cite{chang2023detrdistill}. 
However, according to our observation, Hungarian Matching is not a good choice in IOD, as it cannot effectively preserve the semantic encoding capability of previously-learned queries and fails to mitigate the catastrophic forgetting issue.

As shown in Figure~\ref{query}, we demonstrate why distillation with Hungarian Matching may inevitably suffer from catastrophic forgetting.
Firstly, as shown in Figure~\ref{query}a, we observe that with Hungarian Matching, a certain query (\emph{e.g.,} query \#94 in the figure) in the current-phase model may match different queries of the last-phase model (\emph{e.g.,} query \#55, \#63, \emph{etc.}). 
As a result, the semantic and spatial encoding capability of each query may dramatically change. For example, as illustrated in Figure~\ref{query}b, for the same image, the corresponding semantics and locations of query \#94 change a lot from the last-phase model to the current-phase model (comparing the green and purple bounding boxes).
Consequently, obvious catastrophic forgetting of knowledge encoded in queries still exists.
As illustrated in Figure~\ref{query}c (Left), we observe severe semantic knowledge forgetting. This can be reflected by the total number of queries related to the classification of old categories. A larger total number means more queries encode the semantic information of an old category and thus shows stronger old-category semantic encoding capability of queries. With Hungarian Matching, such a total number obviously decreases from the last-phase model to the current-phase model, exhibiting a clear semantic knowledge forgetting. 
Besides, we also observe slight spatial knowledge forgetting (as illustrated in Figure~\ref{query}c (Middle)). 
Due to both semantic and spatial knowledge forgetting, the detection performance with respect to old categories significantly drops compared to the last-phase model (see Figure~\ref{query}c (Right)). 

In this work, we propose Index-Aligned Query Distillation (IAQD), a novel framework tailored for transformer-based IOD. We aim to preserve the semantic and spatial encoding capability of critical old-category queries, beyond reallocating them via Hungarian Matching, which can effectively mitigate knowledge forgetting. Specifically, IAQD enforces one-to-one correspondence between the queries of the last-phase and current-phase models with the same index. 
Additionally, to avoid disturbing new-category learning, we select queries in the last-phase model that are mainly responsible for detecting old categories (we name these queries as ``proxy query'') to retain knowledge of old categories. 
As a result, only partial queries of the current-phase model are regularized by the queries of the last-phase model, thus making the rest queries more free to learn the knowledge of new categories. This guarantees both the prevention of old-category forgetting and effective learning of new categories. 
Furthermore, to enable more effective exemplar replay, we propose a label realignment strategy to relabel each image with objects from full categories. We directly inherit ground-truth annotations for specific categories, and perform pseudo labeling for the objects of missing categories.
Our method effectively mitigates the catastrophic forgetting issue and achieves state-of-the-art performance on representative benchmarks under different IOD protocols.

In a nutshell, our contributions are summarized as follows:

\begin{itemize}[leftmargin=*]
\item We observe that in transformer-based IOD, utilizing Hungarian Matching to build correspondence between queries of last-phase and current-phase models for distillation is not a good choice, as it may cause the matching to constantly change across different training iterations. 
As a result, the semantic encoding capability of queries with respect to old categories may be largely weakened, and thus the detection performance on old categories drops heavily.

\item We propose IAQD, which performs distillation by aligning queries with the same index to preserve the spatial and semantic encoding capabilities. With the Proxy Query Selection, we only align queries that are crucial for detecting old categories, without constraining the remaining queries. We perform label realignment in Exemplar Replay to further reduce the old knowledge forgetting. 

\item We conduct extensive experiments on the COCO 2017 and PASCAL VOC datasets under different IOD protocols. The experimental results show that our method effectively mitigates the catastrophic forgetting issue and outperforms previous state-of-the-arts. Moreover, ablations verify the effectiveness of each component in IAQD.
\end{itemize}

\section{Related work}
\label{sec:related}

\subsection{Incremental learning}
Incremental learning (also known as continual learning~\cite{aljundi2019task, de2021continual, lopez2017gradient} or lifelong learning~\cite{aljundi2017expert, chaudhry2018efficient, chen2018lifelong}) has become a crucial paradigm in the field of neural networks. Its primary objective is to endow neural networks with the ability to steadily accumulate knowledge while circumventing the issue of catastrophic forgetting~\cite{kemker2018measuring, goodfellow2013empirical, hayes2020remind}. Current approaches to incremental learning can be generally grouped into four main classes. Knowledge distillation (KD) operates by establishing a transfer of knowledge from a pre-trained teacher model to a student model. This is typically achieved by aligning aspects such as logits~\cite{li2017learning, rebuffi2017icarl}, feature maps~\cite{douillard2020podnet}, or other relevant model outputs~\cite{joseph2022energy, liu2023online, pourkeshavarzi2021looking, simon2021learning, tao2020topology, wang2022foster, liu2023continual, wang2025psedet}. Exemplar replay (ER) entails maintaining a small reservoir of samples, known as exemplars, from earlier training rounds~\cite{bang2021rainbow, liu2020mnemonics, prabhu2020gdumb, rebuffi2017icarl, shin2017continual, liu2023continual, kim2024sddgr}. During the training of new tasks, these exemplars are reintroduced and the model can recall and utilize the knowledge gleaned from previous tasks. Regularization methods~\cite{kirkpatrick2017overcoming, zenke2017continual, aljundi2018memory, paik2020overcoming} add penalty terms to the loss function during training to prevent over-updating of parameters crucial for prior tasks. Parameter isolation methods segregate model parameters~\cite{zhang2025dynamic, rusu2016progressive, yoon2019scalable, fernando2017pathnet, kang2022forget} or create task-specific sub-networks for each task~\cite{hung2019compacting, yan2021dynamically, li2019learn, wang2022foster, yoon2017lifelong}. 

\subsection{Incremental object detection}
Incremental Object Detection (IOD) aims to enable object detectors to adapt to new categories while retaining knowledge of old ones. In training data, objects from both old and new categories co-occur, but only the new categories are annotated, which poses a significant challenge. To tackle the issue of catastrophic forgetting, current research mainly relies on two strategies: KD and ER. KD transfers knowledge from the teacher model to the updated one by aligning relevant model outputs or intermediate features. Early works, such as LwF~\cite{li2017learning}, RILOD~\cite{li2019rilod} and iCaRL~\cite{rebuffi2017icarl}, utilize KD to prevent knowledge forgetting. SID~\cite{peng2021sid} proposes Selective and Inter-related Distillation, which conducts distillation at proper locations (non-regression outputs and suitable intermediate layers), and restores the parameters related to old categories on the classification layer for incremental anchor-free object detection. ERD~\cite{feng2022overcoming} proposes Elastic Response Distillation to separately distill knowledge from the classification and regression branches of detectors through an elastic learning approach. ER, on the other hand, preserves old-category knowledge by replaying stored exemplars. \cite{joseph2021towards} use static exemplar sets for fine-tuning, while others introduce adaptive sampling~\cite{liu2020multi} and distribution-aware calibration~\cite{liu2023continual} to improve efficiency and annotation reliability. SDDGR~\cite{kim2024sddgr} synthesizes old-category samples via stable diffusion models to aid the memorization of old knowledge. CL-DETR~\cite{liu2023continual} applies KD and ER to transformer-based incremental object detection. 

\subsection{Transformer-based object detection}
Transformer-based object detection has emerged as a powerful approach, with DETR (DEtection TRansformer)~\cite{carion2020end} leading the way by formulating object detection as a set prediction problem. Its elegant architecture integrates the Transformer with an end-to-end training paradigm, eliminating handcrafted components like anchor generation and NMS. It utilizes a set of learnable object queries and bipartite matching~\cite{sun2021rethinking} to directly predict objects. Deformable DETR~\cite{zhu2020deformable} builds upon DETR by incorporating sparse attention on multi-level feature maps. This not only speeds up the convergence process but also notably improves performance. In addition, there are various DETR variants~\cite{dai2021up, li2022dn, liu2022dab, meng2021conditional,zhang2022dino, zhao2024detrs}. These mainly concentrate on aspects like query refinement and optimization, enhancing the attention mechanism, adding auxiliary loss, as well as improving the training convergence speed and other aspects. These advancements collectively demonstrate the increasing effectiveness and adaptability of transformer-based models within the domain of object detection.
\begin{figure*}[t!]
  \centering
   \includegraphics[width=1.0\linewidth]{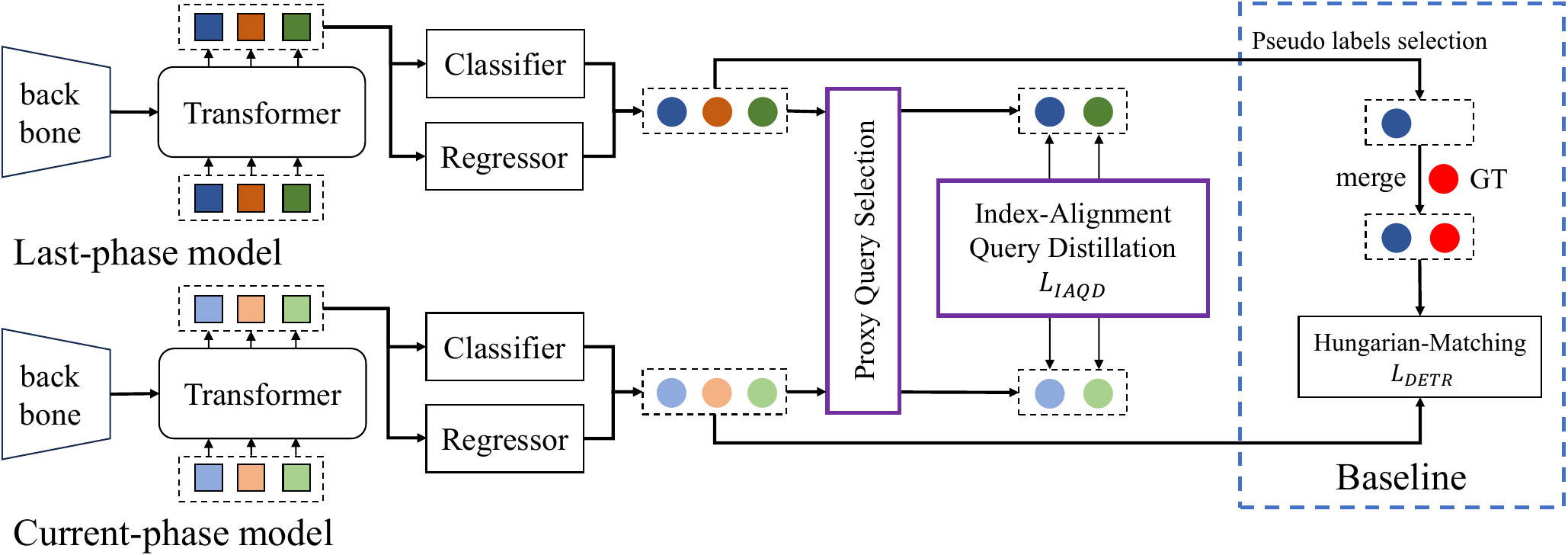}
   \caption{The overview of our proposed method. The baseline method \cite{liu2023continual} only uses the combination of pseudo labels and GT as the supervision signal for $\mathcal{L}_{\text{DETR}}$. Beyond previous work, IAQD introduces two key components. (a) Proxy Query Selection filters queries from the last-phase model that are highly confident in old-category detection (defined by threshold $\tau$). (b) Index-Aligned Query Distillation aligns classifier and regressor outputs of queries with the same index between phases, only distilling old-category knowledge to avoid interfering with new-category learning, whose loss is $\mathcal{L}_{\text{IAQD}}$.}
   \label{framework}
\end{figure*}
\section{Preliminary}
\subsection{DETR}
DETR's architecture comprises a CNN backbone for feature extraction, a transformer encoder-decoder to model global contextual relationships, and a set of learnable object queries. These queries interact with image features via the decoder’s attention mechanism, progressively focusing on potential object locations and semantics, which is achieved by its bipartite matching strategy using the Hungarian algorithm~\cite{kuhn1955hungarian}. This is originally introduced with the concept of set prediction, where a fixed set of queries directly predicts the objects in the image. Given the $N$ object queries and $M$ ground-truth (GT) objects ($M \ll N$), the algorithm assigns each GT (which may be $\varnothing$) to a unique query by minimizing a composite cost, which can be defined as:
\begin{equation}
    \hat{\sigma}=\arg\min_{\sigma}\sum_{i = 1}^{N}\mathcal{L}_{\text{DETR}}(y_i,\hat{y}_{\sigma_i}) ,
\end{equation}
where $\sigma_i$ is the matching mapping from $i$-th GT to the matched query and $\hat{\sigma}$ is the optimal mapping. $y_i=(c_i, b_i)$ represents the $i$-th GT (composed of class $c_i$ and box $b_i$) and $\hat{y}_{\sigma_i}$ is the output of the matched query. $\mathcal{L}_{\text{DETR}}$ is a pair-wise matching cost:
\begin{equation}
    \mathcal{L}_{\text{DETR}}(y_i, \hat{y}_{\sigma_i}) = \mathcal{L}_{\text{cls}}(c_i, \hat{c}_{\sigma_i})+ \mathbbm{1}_{\{c_i \neq \varnothing\}}\mathcal{L}_{\text{loc}}(b_i, \hat{b}_{\sigma_i}),
    \label{detr}
\end{equation}
where $\mathcal{L}_{\text{cls}}$ and $\mathcal{L}_{\text{loc}}$ denote the classification and location losses, respectively. Consequently, each GT is exclusively bound to one query as positive supervision, while all the other queries receive negative training signals. With this optimal assignment strategy, DETR achieves an end-to-end detection framework where queries dynamically specialize in detecting objects through the training.

\subsection{Distillation method in DETR}
\label{Distillation method in DETR}
In particular, the set prediction paradigm of DETR inherently requires alignment between unordered (prediction) sets during training. Consequently, KD in DETR-based incremental learning adopts the same Hungarian Matching strategy to establish correspondence between teacher and student predictions~\cite{chang2023detrdistill} using the distillation loss:
\begin{equation}
    \mathcal{L}_{\text{distill}} = 
    \underbrace{\lambda_{\text{1}} \cdot \mathrm{CE}(p_{\text{T}}^{(i)}, p_{\text{S}}^{(j)})}_{\text{Classification loss}} + 
    \underbrace{(1-\lambda_{\text{1}}) \cdot \mathrm{MSE}(\mathbf{b}_{\text{T}}^{(i)},  \mathbf{b}_{\text{S}}^{(j)})}_{\text{Localization loss}},
    \label{equation_distill}
\end{equation}
where $(i, j)$ represents a matched pair between the predictions of the teacher and student models. The variable $p$ denotes the probabilities, and $\mathbf{b}$ represents the bounding boxes, which are generated by the teacher model with subscript $\text{T}$ and the student model with subscript $\text{S}$ respectively. Additionally, $\lambda_{1}$ is a hyperparameter to balance the classification loss and the localization loss. The first term of Eq.~(\ref{equation_distill}), which calculates the loss using cross entropy (CE), guarantees that the student model can learn the accurate class probabilities from the teacher model. Meanwhile, the second term, which calculates the loss using mean squared error (MSE), enables the student model to precisely match the bounding box predictions. The combination of classification and localization distillation not only effectively transfers knowledge from the teacher model to the student model but also equips the student model with the ability to maintain high-level accuracy in the object detection task.

However, in IOD, Hungarian Matching-based KD will cause the $\hat{\sigma}$ between the teacher model and the student model to change continuously during the training process. This leads to the queries in the student model learning information from the queries of the teacher model with different indexes, resulting in catastrophic forgetting of previous encoded semantic and spatial information. 
\section{Methodology}
In Incremental Object Detection (IOD), the objective involves sequentially training a detection model across multiple phases, with each phase exposing only partial category annotations. Formally, given an image dataset \(\mathcal{D}=\{(x,y)\}\) where \(y\) denotes object annotations containing class labels and the associated bounding boxes, and the category space \(\mathcal{C}=\{1,\ldots, C\}\) is divided into \(T\) disjoint subsets \(\{\mathcal{C}_{t}\}_{t=1}^{T}\). Correspondingly, the training data is also split into \(\mathcal{D}=\mathcal{D}_{1}\cup\cdots\cup\mathcal{D}_{T}\), where each partitioned subset \(\mathcal{D}_{t}\) retains only annotations for categories in \(\mathcal{C}_{t}\) and other object instances remain unlabeled despite their potential presence in images.
During phase \(t\), the current-phase model is only allowed to observe images from \(\mathcal{D}_{t}\) while receiving annotations exclusively for \(\mathcal{C}_{t}\) categories (\( \mathcal{C}_{i} \cap \mathcal{C}_{j} = \varnothing \), \( \forall i \neq j \)). Upon completing phase \(t\), the model transitions to \(\mathcal{D}_{t+1}\) with newly introduced categories \(\mathcal{C}_{t+1}\), maintaining no access to previous annotation sets.

To mitigate catastrophic forgetting, Exemplar Replay mechanism maintains a small portion of historical exemplars through a memory buffer \(\mathcal{E}\). Specifically, the current-phase model preserves samples \(\mathcal{E}_{1:t-1}=\{\mathcal{E}_{1},\cdots,\mathcal{E}_{t-1}\}\) from preceding phases, where \(|\mathcal{E}_{i}|\ll |\mathcal{D}_{i}|\). In this case, the training dataset for phase \(t\) thus becomes \(\mathcal{D}_{t}\cup\mathcal{E}_{1:t-1}\), enabling joint optimization.

\begin{figure}[b!]
  \centering
   \includegraphics[width=0.98\linewidth]{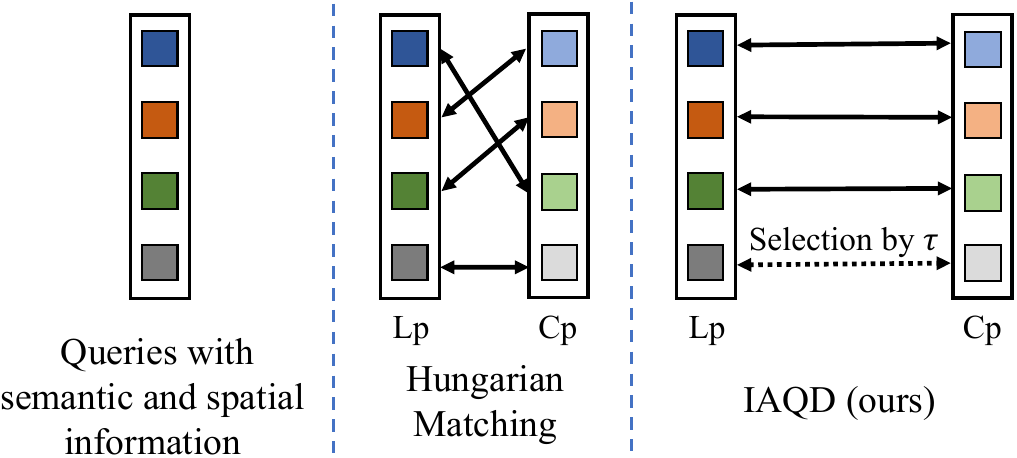}
    \caption{Comparison of our method with Hungarian Matching. Different colors represent query indexes, ``Lp'' and ``Cp'' represent the last-phase and current-phase models, respectively. Middle figure shows that Hungarian Matching enables queries to acquire knowledge from other queries. Right figure demonstrates that queries with the same index are matched. Proxy Query Selection determines whether a query should undergo knowledge distillation.}
   \label{iaqd}
\end{figure}
\subsection{Index-Aligned Query Distillation}
To address the issues raised in Sec.~\ref{Distillation method in DETR}, we propose Index-Aligned Query Distillation (IAQD), which conducts matching of queries based on their index order, as shown in Figure~\ref{framework}. For each query, IAQD maintains consistency between the queries of the last-phase (teacher) and current-phase (student) models, as depicted in Eq.~(\ref{equation_distill}). To distinguish our method, $\mathcal{L}_{\text{distill}}$ is defined as $\mathcal{L}_{\text{IAQD}}$ at this time. The current-phase model is initialized from the last-phase model. Specifically, IAQD aligns the probabilities and bounding boxes obtained from the classifiers and regressors of the two models. We use the CE loss for aligning probabilities and the MSE loss for aligning bounding boxes. Since the teacher lacks new-category semantics, we only align the old-category probabilities ($p_{\text{old}} \in \mathcal{C}_{1:t - 1}$) to prevent semantic interference. Spatial coordinates ($\mathbf{b}_{\text{old}}$) are fully aligned, as box regression is class-agnostic and critical for preserving geometric priors. 
% These queries, trained to cover diverse spatial positions and scales as in~\cite{carion2020end}, inherently span the entire image space. 
By performing distillation through aligning the query indexes, we preserve the semantics of the old categories and the object localization ability of the queries. As shown in Figure~\ref{iaqd}, the matching strategy between teacher queries and student queries is 
\begin{equation}
\sigma_i = i,
\end{equation}
which means the $i$-th teacher query matches $i$-th student query.

\textbf{Proxy Query Selection.}
However, not all queries are equally valuable for object detection. Some queries are closely related to old-category objects, while others may introduce old-category noise or are more suitable for learning new categories. Therefore, we expect to select queries that are important for detecting old categories (termed ``proxy query''), and we refer to this process as Proxy Query Selection. We focus on distilling the proxy queries for which the last-phase model demonstrates high confidence in old-category predictions. As a result, only a portion of the queries in the current-phase model are regularized by those of the last-phase model, allowing the remaining queries to more freely learn the knowledge of new categories. This ensures both the prevention of old-category forgetting and the effective learning of new categories. Mathematically, we define the set $\mathcal{Q}$ as:
\begin{equation}
    \mathcal{Q}=\left\{n \mid \max _{c \in \mathcal{C}_{1: t - 1}} p_{\text{last}}^{(n)}(c) \geq \tau\right\},
    \label{Proxy Query Selection}
\end{equation}
where $p_{\text{last}}^{(n)}(c)$ is the probability of query with index $n$ belonging to old categories $c \in \mathcal{C}_{1: t - 1}$ from the last-phase model. The hyperparameter $\tau$ is a threshold for selecting proxy queries.

\begin{table*}[t!]
    \centering
    % \begin{adjustbox}{width=0.8\textwidth}
        \begin{tabular}{c|lccccccc}
            \toprule
            Setting & Method & Architecture & AP & AP$_{50}$ & AP$_{75}$ & AP$_S$ & AP$_M$ & AP$_L$ \\
            \midrule
            \multirow{7}{*}{40+40}
            & LwF~\cite{li2017learning} & \multirow{4}{*}{GFLv1} & 17.2 & 25.4 & 18.6 & 7.9 & 18.4 & 24.3 \\
            & RILOD~\cite{li2019rilod} & & 29.9 & 45.0 & 32.0 & 15.8 & 33.0 & 40.5 \\
            & SID~\cite{peng2021sid} & & 34.0 & 51.4 & 36.3 & 18.4 & 38.4 & 44.9 \\
            & ERD~\cite{feng2022overcoming} & & 36.9 & 54.5 & 39.6 & 21.3 & 40.4 & 47.5 \\
            \cmidrule(r){2 - 9}
            & CL-DETR~\cite{liu2023continual} & \multirow{4}{*}{Deformable DETR} & 42.0 & 60.1 & 45.9 & 24.0 & 45.3 & 55.6 \\
            & SDDGR~\cite{kim2024sddgr} & & \underline{43.0} & \underline{62.1} & \underline{47.1} & 24.9 & \underline{46.9} & \underline{57.0} \\
            & IAQD w/o ER & & \underline{43.0} & 61.7 & 46.7 & \underline{26.7} & \underline{46.9} & 56.5 \\
            & \textbf{IAQD (ours)} & & \textbf{43.6} & \textbf{62.6} & \textbf{47.7} & \textbf{27.0} & \textbf{47.3} & \textbf{57.3} \\
            \midrule
            \multirow{7}{*}{70+10} & LwF~\cite{li2017learning} & \multirow{4}{*}{GFLv1} & 7.1 & 12.4 & 7.0 & 4.8 & 9.5 & 10.0 \\
            & RILOD~\cite{li2019rilod} & & 24.5 & 37.9 & 25.7 & 14.2 & 27.4 & 33.5 \\
            & SID~\cite{peng2021sid} & & 32.8 & 49.0 & 35.0 & 17.1 & 36.9 & 44.5 \\
            & ERD~\cite{feng2022overcoming} & & 34.9 & 51.9 & 37.4 & 18.7 & 38.8 & 45.5  \\
            \cmidrule(lr){2 - 9}
            & CL-DETR~\cite{liu2023continual} & \multirow{4}{*}{Deformable DETR} & 40.4 & 58.0 & 43.9 & 23.8 & 43.6 & 53.5 \\
            & SDDGR~\cite{kim2024sddgr} & & 40.9 & 59.5 & 44.8 & \underline{23.9} & 44.7 & 54.0 \\
            & IAQD w/o ER & & \underline{41.8} & \underline{60.4} & \underline{45.5} & \textbf{25.5} & \underline{45.4} & \underline{54.8} \\
            & \textbf{IAQD (ours)} & & \textbf{43.2} & \textbf{62.2} & \textbf{46.9} & \textbf{25.5} & \textbf{46.5} & \textbf{56.8} \\
        \bottomrule
        \end{tabular}
    % \end{adjustbox}
    \caption{IOD results (\%) on COCO 2017 dataset under  \textit{Protocol A}. The numbers for previous works are cited from SDDGR~\cite{kim2024sddgr}. The best performance is highlighted \textbf{in bold} and second best is \underline{underline}.}
    \label{traditional protocol}
\end{table*}
\begin{table*}[th!]
    \centering
    % \begin{adjustbox}{width=0.8\textwidth}
        \begin{tabular}{c|lccccccc}
            \toprule
            Setting & Method & Architecture & AP & AP$_{50}$ & AP$_{75}$ & AP$_S$ & AP$_M$ & AP$_L$ \\
            \midrule
            \multirow{5}{*}{40+40} & LwF~\cite{li2017learning} & \multirow{6}{*}{Deformable DETR} & 23.9 & 41.5 & 25.0 & 12.0 & 26.4 & 33.0 \\
            & iCaRL~\cite{rebuffi2017icarl} & & 33.4 & 52.0 & 36.0 & 18.0 & 36.4 & 45.5 \\
            & ERD~\cite{feng2022overcoming} & & 36.0 & 55.2 & 38.7 & 19.5 & 38.7 & 49.0 \\
            & CL-DETR~\cite{liu2023continual} & & 37.5 & 55.1 & 40.3 & 20.9 & 40.8 & 50.7 \\
            & IAQD w/o ER & & \underline{38.9} & \underline{56.7} & \underline{42.2} & \underline{22.0} & \underline{42.0} & \underline{51.3} \\
            & \textbf{IAQD (ours)} & & \textbf{39.9} & \textbf{57.7} & \textbf{43.4} & \textbf{22.9} & \textbf{43.0} & \textbf{52.2} \\
            \midrule
            \multirow{5}{*}{70+10}
            & LwF~\cite{li2017learning} & \multirow{6}{*}{Deformable DETR} & 24.5 & 36.6 & 26.7 & 12.4 & 28.2 & 35.2 \\
            & iCaRL~\cite{rebuffi2017icarl} & & 35.9 & 52.5 & 39.2 & 19.1 & 39.4 & 48.6 \\
            & ERD~\cite{feng2022overcoming} & & 36.9 & 55.7 & 40.1 & 21.4 & 39.6 & 48.7 \\
            & CL-DETR~\cite{liu2023continual} & & 40.1 & 57.8 & 43.7 & 23.2 & 43.2 & 52.1 \\
            & IAQD w/o ER & & \underline{40.9} & \underline{59.5} & \underline{44.4} & \underline{24.1} & \underline{44.3} & \underline{53.6} \\
            & \textbf{IAQD (ours)} & & \textbf{42.2} & \textbf{60.5} & \textbf{46.2} & \textbf{25.1} & \textbf{45.2} & \textbf{55.6} \\
        \bottomrule
        \end{tabular}
    % \end{adjustbox}
    \caption{IOD results (\%) on COCO 2017 dataset under \textit{Protocol B}. The numbers for previous works are cited from CL-DETR~\cite{liu2023continual}.}
    \label{revised protocol}
\end{table*}

\subsection{Overall Loss}
During the training of IOD, the dataset $\mathcal{D}_{t}$ lacks labels for old categories $\mathcal{C}_{1:t - 1}$. To address this issue, we let the last-phase model generate pseudo labels $\hat{y}_{\text{old}}$ for old categories on $\mathcal{D}_{t}$ by setting a confidence threshold. These hard labels, consisting of category IDs and bounding boxes, are merged with the GT annotations of new categories $\mathcal{C}_{t}$, and the combined annotations serve as $y$ in Eq.~(\ref{detr}). Subsequently, the current-phase model can be trained through the standard training process of DETR. The total training loss combines both Hungarian Matching and IAQD:
\begin{align}
\mathcal{L}_{\text{total}}&=\mathcal{L}_{\text{DETR}}+\lambda_{2}\cdot\mathcal{L}_{\text{IAQD}},
\end{align}
where $\lambda_{2}$ is a hyperparameter to balance the loss terms and $\mathcal{L}_{\text{IAQD}}$ is defined as:
\begin{equation}
    \begin{split}
        \mathcal{L}_{\text{IAQD}} = & 
        \lambda_{\text{1}} \cdot \mathrm{CE}(p_{\text{last}}^{(i)}, \tilde{p}_{\text{cur}}^{(i)}) \\
        & + 
        (1-\lambda_{\text{1}}) \cdot \mathrm{MSE}(\mathbf{b}_{\text{last}}^{(i)}, \mathbf{b}_{\text{cur}}^{(i)}),\forall i \in \mathcal{Q},
    \end{split}   
    \label{equation_iaqd}
\end{equation}
where $(p_\text{last}, b_\text{last})$ are the probabilities and bounding boxes output by the last-phase model. $\tilde{p}_{\text{cur}}$ is the sub-vector of the probabilities $p_\text{cur}$ output by the current-phase model, which refers to the probabilities of the old categories, and $\mathbf{b}_{\text{cur}}$ are the bounding boxes output by the current-phase model.

\subsection{Exemplar Replay with Label Realignment}
Following CL-DETR~\cite{liu2023continual}, after the current phase training, we also conduct an exemplar replay process. In the Exemplar Replay stage of the current phase, there may be cases where old exemplars $\mathcal{E}_{1:t-1}$ lack annotations for new categories $\mathcal{C}_{t}$, and the new data $\mathcal{D}_{t}$ also lacks annotations for old categories $\mathcal{C}_{1:t-1}$. To address this issue, we propose the Exemplar Replay with Label Realignment. It uses the current-phase model to generate pseudo labels for these categories that have missing annotations in the exemplars, and we retain the GT annotations. This enriches both the old and new data, making the label information more comprehensive. Moreover, during the ER stage, $\mathcal{E}_{1:t - 1}$ has annotations of old categories, which helps to mitigate knowledge forgetting. We choose not to employ $\mathcal{L}_{\text{IAQD}}$ at this stage, which also reduces computational costs. The loss in this stage is merely $\mathcal{L}_{\text{DETR}}$.
\begin{table*}[t!]
    \centering
    % \begin{adjustbox}{width=0.8\textwidth}
        \begin{tabular}{cccccccc}
            \toprule
            \multirow{2}{*}{Method} & \multirow{2}{*}{Architecture} & \multicolumn{4}{c}{$40+10+10+10+10$} & \multicolumn{2}{c}{$40+20+20$} \\
            \cmidrule(lr){3 - 6} \cmidrule(lr){7 - 8}
            & & $(40-50)$ & $(50-60)$ & $(60-70)$ & $(70-80)$ & $(40-60)$ & $(60-80)$ \\
            \midrule
            RILOD~\cite{li2019rilod} & \multirow{3}{*}{GFLv1} & 25.4 & 11.2 & 10.5 & 8.4 & 27.8 & 15.8 \\
            SID~\cite{peng2021sid} & & 34.6 & 24.1 & 14.6 & 12.6 & 34.0 & 23.8 \\
            ERD~\cite{feng2022overcoming} & & 36.4 & 30.8 & 26.2 & 20.7 & 36.7 & 32.4 \\
            \midrule
            CL-DETR~\cite{liu2023continual} & \multirow{3}{*}{Deformable DETR} & - & - & - & 28.1 & - & 35.3 \\
            SDDGR~\cite{kim2024sddgr} & & 42.3 & 40.6 & 40.0 & 36.8 & 42.5 & 41.1 \\
            \textbf{IAQD (ours)} & & \textbf{44.8} & \textbf{43.3} & \textbf{42.2} & \textbf{40.5} & \textbf{44.8} & \textbf{42.8} \\
            \bottomrule
        \end{tabular}
    % \end{adjustbox}
    \caption{IOD results (AP, \%) on COCO 2017 dataset under the \textit{Protocol A} in the multi-phase setting. In the first phase, normal training is conducted with 40 categories, followed by the addition of 10 and 20 new categories in the following 4-phase and 2-phase settings each time, respectively. The numbers in previous works are cited from SDDGR~\cite{kim2024sddgr}.}
    \label{multi-phase}
\end{table*}
\section{Experiments}
\subsection{Setup}
\textbf{Datasets and Metrics.} To enable a fair and comprehensive comparison with previous works, we conduct experiments on the COCO 2017 dataset~\cite{lin2014microsoft}, which contains 80 object categories. To better demonstrate generalizability, we perform ablation studies on the PASCAL VOC dataset~\cite{everingham2010pascal}, which covers 20 foreground classes. For performance evaluation, we employ standard COCO metrics, 
\textbf{AP} refers to the average precision calculated across IoU thresholds ranging from 0.5 to 0.95. Specifically, \textbf{AP$_{50}$} and \textbf{AP$_{75}$} are evaluated at IoU thresholds of 0.5 and 0.75, respectively. Additionally, there are scale-specific metrics: \textbf{AP$_S$}, \textbf{AP$_M$}, and \textbf{AP$_L$} assess small, medium, and large objects, respectively. Here, small objects are defined as those with a pixel area of less than $32^2$, medium objects have a pixel area between $32^2$(inclusive) and $96^2$(exclusive), and large objects have a pixel area of $96^2$ or greater. \textbf{AP} is the evaluation metric for COCO, and \textbf{AP$_{50}$} is used for VOC.

\textbf{Protocols.} We evaluate our framework under two distinct incremental learning protocols: (1) \textit{Protocol A}~\cite{shmelkov2017incremental}: At each phase $\tau$, the model processes all images containing at least one object belonging to the current category subset $\mathcal{C}_{\tau}$. This often results in image overlap across different phases. Specifically, an image can be part of multiple phases if it contains objects from different category subsets. 
(2) \textit{Protocol B}~\cite{liu2023continual}: Instead of partitioning the dataset based on current categories, we directly split the entire dataset. This approach effectively ensures that the model encounters non-overlapping images across different phases. However, it may lead to situations where images in a particular phase do not contain any objects belonging to the current category subset.

\textbf{Implementation Details.} We adopt Deformable DETR with a ResNet-50 backbone pre-trained on ImageNet~\cite{zhu2020deformable}, disabling iterative bounding box refinement and the two-stage mechanism to align with standard IOD protocols. The initial model is trained for 50 epochs with a batch size of 2 per GPU (4 GPUs in total). For each new phase, we use the AdamW optimizer~\cite{loshchilov2017decoupled} to train the model for 50 epochs in the incremental stage and perform fine-tuning for 20 epochs in the ER stage (learning rate: $6 \times 10^{-5}$, weight decay: $1\times 10^{-4}$). To ensure comparability and reproducibility, we randomize the category order following the predefined random seed and category order in CL-DETR~\cite{liu2023continual}. Additionally, we repeat the experiments three times with different random seeds and take the average value as the final performance. We adopt both two-phase and multiple-phase settings, and also set the total memory budget for the exemplars to be 10\% of the total dataset size. In accordance with the method presented in~\cite{liu2023continual}, we incorporate 10\% of the new-category data along with exemplars for ER stage fine-tuning. The threshold for selecting pseudo labels in the incremental stage of the current-phase is set to 0.4, while the threshold in the ER stage is set to 0.6. The hyperparameters are set as $\tau = 0.1$, $\lambda_1 = 0.5$, $\lambda_2 = 1$. All experiments are conducted on 4 NVIDIA 4090 GPUs.

\subsection{Comparison with previous state-of-the-arts}
\textbf{Two-phase setting:} Table~\ref{traditional protocol} presents the performance of IAQD and other methods on the COCO 2017 dataset under the \textit{Protocol A}. We have three observations. 
1) Overall, the performance of the Deformable Detr-based methods is better than that of the GFLv1-based methods. 
2) In both settings, IAQD outperforms state-of-the-art approaches. Specifically, in the $70 + 10$ setting, IAQD achieves a 2.3\% improvement in AP compared to the previous best performing method SDDGR~\cite{kim2024sddgr}. IAQD proves more effective in the more challenging $70 + 10$ setting, where category imbalance is more severe.
3) Moreover, across the two settings, the difference in AP of IAQD is only 0.4\%. In contrast, for SDDGR, this difference is 2.1\%. This shows that IAQD is less affected by different IOD settings. 
And under the \textit{Protocol B}, as shown in Table~\ref{revised protocol}, our method achieves AP of 39.9\% and 42.2\% in the $40 + 40$ and $70 + 10$ settings, respectively, surpassing CL-DETR's performance by 2.4\% and 2.1\% under each setting. 
In addition, the performance of IAQD without ER almost always compares favourably against that of the previous SOTA methods.
This demonstrates that IAQD without ER can still maintain superior performance.

\textbf{Multi-phase setting:} 
Table~\ref{multi-phase} shows the performance on COCO 2017 dataset under the \textit{Protocol A} in the $40 + 20\times2$ and $40 + 10\times4$ settings. Here, for example, $40 + 20\times2$ indicates that 40 categories are learned in the first phase, and 20 categories are learned in each of the next two phases. These multi-phase settings are designed to evaluate the long-term incremental learning capabilities, posing a significant challenge as the increasing number of training phases intensifies the forgetting problem. In comparison to SDDGR, IAQD demonstrates remarkable superiority, particularly in the $40 + 10\times4$ setting with an improvement of 3.7\%. This shows that as the complexity of the incremental learning task rises, IAQD can effectively mitigate the forgetting issue, enabling it to retain more knowledge acquired from previous phases.

\subsection{Ablations}
\textbf{Effect of each component}. Table~\ref{component} details the systematic ablation of our framework's components under the \textit{Protocol A} in the $70 + 10$ setting. We incrementally added each component, and the performance improved at each step, demonstrating that each of our designs is effective. The application of Index-Aligned Query Distillation enhances the performance of old categories by 0.7\% but slightly impairs the performance of new categories. Next, with the introduction of Proxy Query Selection, the performance of new categories is improved by 0.6\% without compromising that of old categories. This demonstrates that we retain queries crucial for detecting old categories while allowing other queries to learn new categories, thereby achieving a relatively good performance balance between old and new categories. Finally, ER with Label Realignment further boosts the performance of old categories by 1.5\%, showing a further mitigation of the forgetting of old knowledge.
\begin{table}[h]
    \centering
    \small
    % \begin{adjustbox}{width=0.98\linewidth}
        \begin{tabular}{lccc}
            \toprule
            Method & All & Old & New \\
            \midrule
            Baseline & 40.8 & 41.1 & \textbf{45.2} \\
            +Index-Aligned Query Distillation & 41.5 & 41.9 & 43.8 \\
            ++Proxy Query Selection & 41.7 & 42.1 & 44.4 \\
            +++ER with Label Realignment & \textbf{43.2} & \textbf{43.6} & 43.5 \\
            \bottomrule
        \end{tabular}
    % \end{adjustbox}
    \caption{Effect (AP, \%) of each component under the \textit{Protocol A} in the $70 + 10$ setting, where ``All'', ``Old'', and ``New'' represent the performance in all categories, old categories, and new categories, respectively. For the baseline, we use a 0.4 threshold as the pseudo-label selection method for CL-DETR.}
    \label{component}
\end{table}
\begin{figure*}[h]
    \centering
    \includegraphics[width=1.0\linewidth]{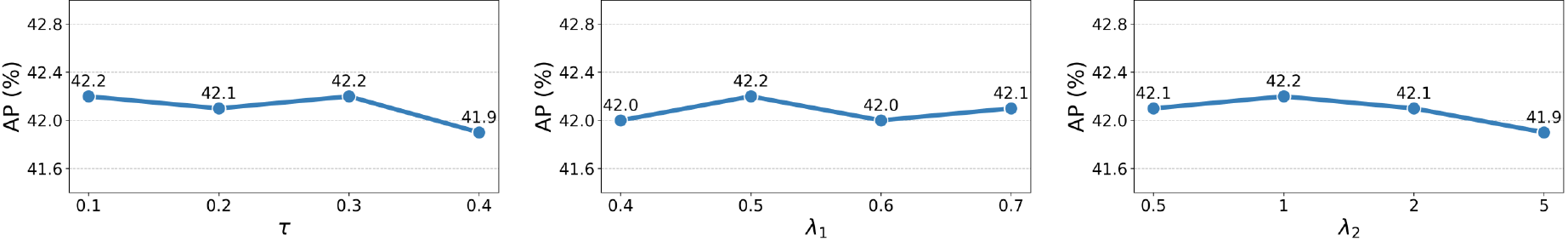}
    \caption{Sensitivity (AP, \%) to $\tau$, $\lambda_1$ and $\lambda_2$ under the \textit{Protocol B} in the $70+10$ setting.}
    \label{sensitivity}
\end{figure*}

\textbf{Generalize to other transformer-based model}. We apply IAQD and CL-DETR to DINO-DETR, each trained for 12 epochs in the $70 + 10$ setting, followed by 5 epochs of exemplar replay. 
Table~\ref{dino} shows IAQD consistently outperforms CL-DETR under both \textit{Protocol A} and \textit{B}. These improvements demonstrate that IAQD can be readily applied to other transformer-based detectors without additional modifications, achieving favourable performance.
\begin{table}[h!]
    \centering
    \begin{adjustbox}{width=1.0\linewidth}
        \begin{tabular}{c|lccccc}
            \toprule
            Protocol & Method & Architecture & AP & AP$_{50}$ & AP$_{75}$ \\
            \midrule
            \multirow{2}{*}{A} & CL-DETR~\cite{liu2023continual} & \multirow{4}{*}{DINO} & 45.9 & 62.1 & 50.1 \\
            & \textbf{IAQD (ours)} & & \textbf{47.4} & \textbf{63.9} & \textbf{51.8} \\
            \cmidrule(lr){1 - 2} \cmidrule(lr){4 - 6}
            \multirow{2}{*}{B} & CL-DETR~\cite{liu2023continual} & & 45.8 & 62.4 & 49.9 \\
            & \textbf{IAQD (ours)} & & \textbf{46.4} & \textbf{62.9} & \textbf{50.8} \\
        \bottomrule
        \end{tabular}
    \end{adjustbox}
    \caption{IOD results (\%) on COCO 2017 dataset in $70+10$ setting.}
    \label{dino}
\end{table}

\textbf{Extend to other benchmark}. As shown in Table~\ref{voc}, we apply IAQD and CL-DETR to the PASCAL VOC dataset. The 20 categories are divided into two settings: $10+10$ and $15+5$. Under both \textit{Protocol A} and \textit{B}, IAQD consistently outperforms CL-DETR, highlighting its consistent generalization ability across diverse datasets.
\begin{table}[h!]
    \centering
    \small
    % \begin{adjustbox}{width=0.96\linewidth}
        \begin{tabular}{c|clccc}
            \toprule
            Protocol & Setting & Method & AP & AP$_{50}$ & AP$_{75}$ \\
            \midrule
            \multirow{4}{*}{A} & \multirow{2}{*}{10 + 10} & CL-DETR & 44.1 & 67.8 & 47.8 \\
            & & \textbf{IAQD} & \textbf{50.0} & \textbf{74.7} & \textbf{55.2} \\
            \cmidrule(lr){2 - 6}
            & \multirow{2}{*}{15 + 5}
            & CL-DETR & 47.5 & 71.3 & 52.0 \\
            & & \textbf{IAQD} & \textbf{50.1} & \textbf{75.0} & \textbf{55.2} \\
            \midrule            
            \multirow{4}{*}{B} & \multirow{2}{*}{10 + 10} & CL-DETR & 34.0 & 56.9 & 35.5 \\
            & & \textbf{IAQD} & \textbf{45.3} & \textbf{71.7} & \textbf{48.5} \\
            \cmidrule(lr){2 - 6}
            & \multirow{2}{*}{15 + 5}
            & CL-DETR & 42.1 & 66.3 & 45.3 \\
            & & \textbf{IAQD} & \textbf{47.8} & \textbf{73.3} & \textbf{52.5} \\
        \bottomrule
        \end{tabular}
    % \end{adjustbox}
    \caption{IOD results (\%) on PASCAL VOC dataset.}
    \label{voc}
\end{table}

\textbf{Sensitivity to hyperparameters}. As shown in Figure~\ref{sensitivity}, we conducted sensitivity analyses to the Proxy Query Selection threshold $\tau$ and the loss calculation weight parameters $\lambda_1$ and $\lambda_2$ under the \textit{Protocol B} in the $70+10$ setting. The results demonstrate that our model achieves optimal performance when the query threshold $\tau$ is 0.1. Even as $\tau$ increases from 0.1 to 0.4, performance only exhibits minor fluctuations. Additionally, the model reaches its best performance when $\lambda_1$ is 0.5 and $\lambda_2$ is 1. When $\lambda_1$ varies between 0.3 and 0.7, and $\lambda_2$ between 0.5 and 5, the model maintains relatively stable detection performance. This demonstrates that our method can tolerate a certain range of hyperparameter variations with relatively limited performance degradation.
\begin{figure*}[t!]
    \centering
    \includegraphics[width=1.0\linewidth]{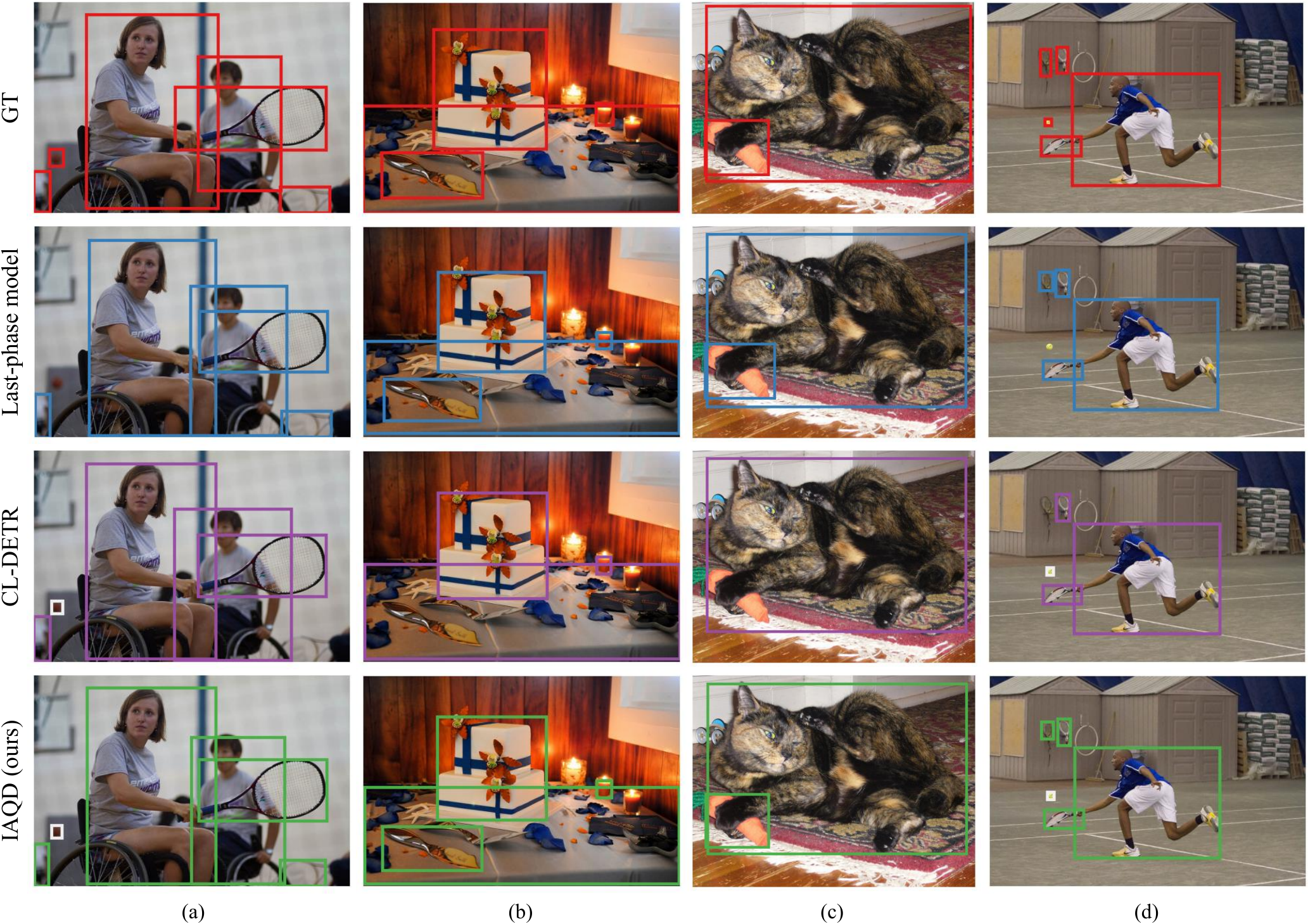}
    \caption{Visualization the results of object detection on COCO 2017 dataset in the $70 + 10$ setting. The white bounding boxes enclose the new categories, while those of other colors represent the detection results of different models. It can be seen that our method is able to detect the objects that are detected by the last-phase model in CL-DETR~\cite{liu2023continual} but missed by CL-DETR’s current-phase model.}
    \label{visualization}
\end{figure*}

\textbf{Visualizations}. Figure~\ref{visualization}  visualizes the detection performance of ground truth (GT) and different models under the \textit{Protocol B} in the $70 + 10$ setting. In the figure, objects framed by white bounding boxes belong to the new categories introduced in current phase, while the rest represent the old categories from last phase. It’s evident that some GT objects can be successfully recognized by the CL-DETR~\cite{liu2023continual} last-phase model. However, after the introduction of new categories, the CL-DETR model~\cite{liu2023continual} fails to recognize these objects. In contrast, our proposed IAQD effectively retains the detection ability of the last-phase model for old category objects. Moreover, IAQD demonstrates excellent performance in recognizing new objects, thus achieving a balance between the detection of old and new objects.
\section{Conclusion}
This paper introduces IAQD, a novel transformer-based IOD method. IAQD retains old knowledge through a one-to-one distillation method based on query index. Meanwhile, we adopt Proxy Query Selection to select queries that are important for old knowledge. Additionally, IAQD strengthens Exemplar Replay via label realignment. Extensive experiments on representative benchmarks under different IOD protocols show our method outperforms previous state-of-the-arts remarkably. In the future, we plan to extend our method to other incremental learning scenarios.
\section*{Acknowledgments}
This research project is supported by National Key R\&D Program of China under Grant 2022YFB3104900.
{
    \small
    \bibliographystyle{plain}
    \bibliography{main}

\begin{thebibliography}{10}

\bibitem{aljundi2018memory}
Rahaf Aljundi, Francesca Babiloni, Mohamed Elhoseiny, Marcus Rohrbach, and Tinne Tuytelaars.
\newblock Memory aware synapses: Learning what (not) to forget.
\newblock In {\em Proceedings of the European conference on computer vision (ECCV)}, pages 139--154, 2018.

\bibitem{aljundi2017expert}
Rahaf Aljundi, Punarjay Chakravarty, and Tinne Tuytelaars.
\newblock Expert gate: Lifelong learning with a network of experts.
\newblock In {\em Proceedings of the IEEE conference on computer vision and pattern recognition}, pages 3366--3375, 2017.

\bibitem{aljundi2019task}
Rahaf Aljundi, Klaas Kelchtermans, and Tinne Tuytelaars.
\newblock Task-free continual learning.
\newblock In {\em Proceedings of the IEEE/CVF conference on computer vision and pattern recognition}, pages 11254--11263, 2019.

\bibitem{bai2024revisiting}
Liang Bai, Hong Song, Tao Feng, Tianyu Fu, Qingzhe Yu, and Jian Yang.
\newblock Revisiting class-incremental object detection: An efficient approach via intrinsic characteristics alignment and task decoupling.
\newblock {\em Expert Systems with Applications}, 257:125057, 2024.

\bibitem{bai2025class}
Liang Bai, Hangjie Yuan, Hong Song, Tao Feng, and Jian Yang.
\newblock Class-incremental player detection with refined response-based knowledge distillation.
\newblock {\em IEEE Transactions on Instrumentation and Measurement}, 2025.

\bibitem{bang2021rainbow}
Jihwan Bang, Heesu Kim, YoungJoon Yoo, Jung-Woo Ha, and Jonghyun Choi.
\newblock Rainbow memory: Continual learning with a memory of diverse samples.
\newblock In {\em Proceedings of the IEEE/CVF conference on computer vision and pattern recognition}, pages 8218--8227, 2021.

\bibitem{caccia2021new}
Lucas Caccia, Rahaf Aljundi, Nader Asadi, Tinne Tuytelaars, Joelle Pineau, and Eugene Belilovsky.
\newblock New insights on reducing abrupt representation change in online continual learning.
\newblock {\em arXiv preprint arXiv:2104.05025}, 2021.

\bibitem{carion2020end}
Nicolas Carion, Francisco Massa, Gabriel Synnaeve, Nicolas Usunier, Alexander Kirillov, and Sergey Zagoruyko.
\newblock End-to-end object detection with transformers.
\newblock In {\em European conference on computer vision}, pages 213--229. Springer, 2020.

\bibitem{chang2023detrdistill}
Jiahao Chang, Shuo Wang, Hai-Ming Xu, Zehui Chen, Chenhongyi Yang, and Feng Zhao.
\newblock Detrdistill: A universal knowledge distillation framework for detr-families.
\newblock In {\em Proceedings of the IEEE/CVF international conference on computer vision}, pages 6898--6908, 2023.

\bibitem{chaudhry2018efficient}
Arslan Chaudhry, Marc'Aurelio Ranzato, Marcus Rohrbach, and Mohamed Elhoseiny.
\newblock Efficient lifelong learning with a-gem.
\newblock {\em arXiv preprint arXiv:1812.00420}, 2018.

\bibitem{chen2018lifelong}
Zhiyuan Chen and Bing Liu.
\newblock {\em Lifelong machine learning}.
\newblock Morgan \& Claypool Publishers, 2018.

\bibitem{dai2021up}
Zhigang Dai, Bolun Cai, Yugeng Lin, and Junying Chen.
\newblock Up-detr: Unsupervised pre-training for object detection with transformers.
\newblock In {\em Proceedings of the IEEE/CVF conference on computer vision and pattern recognition}, pages 1601--1610, 2021.

\bibitem{de2021continual}
Matthias De~Lange, Rahaf Aljundi, Marc Masana, Sarah Parisot, Xu~Jia, Ale{\v{s}} Leonardis, Gregory Slabaugh, and Tinne Tuytelaars.
\newblock A continual learning survey: Defying forgetting in classification tasks.
\newblock {\em IEEE transactions on pattern analysis and machine intelligence}, 44(7):3366--3385, 2021.

\bibitem{douillard2020podnet}
Arthur Douillard, Matthieu Cord, Charles Ollion, Thomas Robert, and Eduardo Valle.
\newblock Podnet: Pooled outputs distillation for small-tasks incremental learning.
\newblock In {\em Computer vision--ECCV 2020: 16th European conference, Glasgow, UK, August 23--28, 2020, proceedings, part XX 16}, pages 86--102. Springer, 2020.

\bibitem{everingham2010pascal}
Mark Everingham, Luc Van~Gool, Christopher~KI Williams, John Winn, and Andrew Zisserman.
\newblock The pascal visual object classes (voc) challenge.
\newblock {\em International journal of computer vision}, 88(2):303--338, 2010.

\bibitem{feng2022overcoming}
Tao Feng, Mang Wang, and Hangjie Yuan.
\newblock Overcoming catastrophic forgetting in incremental object detection via elastic response distillation.
\newblock In {\em Proceedings of the IEEE/CVF Conference on Computer Vision and Pattern Recognition}, pages 9427--9436, 2022.

\bibitem{fernando2017pathnet}
Chrisantha Fernando, Dylan Banarse, Charles Blundell, Yori Zwols, David Ha, Andrei~A Rusu, Alexander Pritzel, and Daan Wierstra.
\newblock Pathnet: Evolution channels gradient descent in super neural networks.
\newblock {\em arXiv preprint arXiv:1701.08734}, 2017.

\bibitem{goodfellow2013empirical}
Ian~J Goodfellow, Mehdi Mirza, Da~Xiao, Aaron Courville, and Yoshua Bengio.
\newblock An empirical investigation of catastrophic forgetting in gradient-based neural networks.
\newblock {\em arXiv preprint arXiv:1312.6211}, 2013.

\bibitem{hayes2020remind}
Tyler~L Hayes, Kushal Kafle, Robik Shrestha, Manoj Acharya, and Christopher Kanan.
\newblock Remind your neural network to prevent catastrophic forgetting.
\newblock In {\em European conference on computer vision}, pages 466--483. Springer, 2020.

\bibitem{hinton2015distilling}
Geoffrey Hinton, Oriol Vinyals, and Jeff Dean.
\newblock Distilling the knowledge in a neural network.
\newblock {\em arXiv preprint arXiv:1503.02531}, 2015.

\bibitem{hou2019learning}
Saihui Hou, Xinyu Pan, Chen~Change Loy, Zilei Wang, and Dahua Lin.
\newblock Learning a unified classifier incrementally via rebalancing.
\newblock In {\em Proceedings of the IEEE/CVF conference on computer vision and pattern recognition}, pages 831--839, 2019.

\bibitem{hu2021distilling}
Xinting Hu, Kaihua Tang, Chunyan Miao, Xian-Sheng Hua, and Hanwang Zhang.
\newblock Distilling causal effect of data in class-incremental learning.
\newblock In {\em Proceedings of the IEEE/CVF conference on Computer Vision and Pattern Recognition}, pages 3957--3966, 2021.

\bibitem{hung2019compacting}
Ching-Yi Hung, Cheng-Hao Tu, Cheng-En Wu, Chien-Hung Chen, Yi-Ming Chan, and Chu-Song Chen.
\newblock Compacting, picking and growing for unforgetting continual learning.
\newblock {\em Advances in neural information processing systems}, 32, 2019.

\bibitem{jiang2022review}
Peiyuan Jiang, Daji Ergu, Fangyao Liu, Ying Cai, and Bo~Ma.
\newblock A review of yolo algorithm developments.
\newblock {\em Procedia computer science}, 199:1066--1073, 2022.

\bibitem{joseph2022energy}
KJ~Joseph, Salman Khan, Fahad~Shahbaz Khan, Rao~Muhammad Anwer, and Vineeth~N Balasubramanian.
\newblock Energy-based latent aligner for incremental learning.
\newblock In {\em Proceedings of the IEEE/CVF Conference on Computer Vision and Pattern Recognition}, pages 7452--7461, 2022.

\bibitem{joseph2021towards}
KJ~Joseph, Salman Khan, Fahad~Shahbaz Khan, and Vineeth~N Balasubramanian.
\newblock Towards open world object detection.
\newblock In {\em Proceedings of the IEEE/CVF conference on computer vision and pattern recognition}, pages 5830--5840, 2021.

\bibitem{kang2022forget}
Haeyong Kang, Rusty John~Lloyd Mina, Sultan Rizky~Hikmawan Madjid, Jaehong Yoon, Mark Hasegawa-Johnson, Sung~Ju Hwang, and Chang~D Yoo.
\newblock Forget-free continual learning with winning subnetworks.
\newblock In {\em International Conference on Machine Learning}, pages 10734--10750. PMLR, 2022.

\bibitem{kang2023alleviating}
Mengxue Kang, Jinpeng Zhang, Jinming Zhang, Xiashuang Wang, Yang Chen, Zhe Ma, and Xuhui Huang.
\newblock Alleviating catastrophic forgetting of incremental object detection via within-class and between-class knowledge distillation.
\newblock In {\em Proceedings of the IEEE/CVF International Conference on Computer Vision}, pages 18894--18904, 2023.

\bibitem{kemker2018measuring}
Ronald Kemker, Marc McClure, Angelina Abitino, Tyler Hayes, and Christopher Kanan.
\newblock Measuring catastrophic forgetting in neural networks.
\newblock In {\em Proceedings of the AAAI conference on artificial intelligence}, volume~32, 2018.

\bibitem{kim2024sddgr}
Junsu Kim, Hoseong Cho, Jihyeon Kim, Yihalem~Yimolal Tiruneh, and Seungryul Baek.
\newblock Sddgr: Stable diffusion-based deep generative replay for class incremental object detection.
\newblock In {\em Proceedings of the IEEE/CVF Conference on Computer Vision and Pattern Recognition}, pages 28772--28781, 2024.

\bibitem{kirkpatrick2017overcoming}
James Kirkpatrick, Razvan Pascanu, Neil Rabinowitz, Joel Veness, Guillaume Desjardins, Andrei~A Rusu, Kieran Milan, John Quan, Tiago Ramalho, Agnieszka Grabska-Barwinska, et~al.
\newblock Overcoming catastrophic forgetting in neural networks.
\newblock {\em Proceedings of the national academy of sciences}, 114(13):3521--3526, 2017.

\bibitem{kuhn1955hungarian}
Harold~W Kuhn.
\newblock The hungarian method for the assignment problem.
\newblock {\em Naval research logistics quarterly}, 2(1-2):83--97, 1955.

\bibitem{li2019rilod}
Dawei Li, Serafettin Tasci, Shalini Ghosh, Jingwen Zhu, Junting Zhang, and Larry Heck.
\newblock Rilod: Near real-time incremental learning for object detection at the edge.
\newblock In {\em Proceedings of the 4th ACM/IEEE Symposium on Edge Computing}, pages 113--126, 2019.

\bibitem{li2022dn}
Feng Li, Hao Zhang, Shilong Liu, Jian Guo, Lionel~M Ni, and Lei Zhang.
\newblock Dn-detr: Accelerate detr training by introducing query denoising.
\newblock In {\em Proceedings of the IEEE/CVF conference on computer vision and pattern recognition}, pages 13619--13627, 2022.

\bibitem{li2019learn}
Xilai Li, Yingbo Zhou, Tianfu Wu, Richard Socher, and Caiming Xiong.
\newblock Learn to grow: A continual structure learning framework for overcoming catastrophic forgetting.
\newblock In {\em International conference on machine learning}, pages 3925--3934. PMLR, 2019.

\bibitem{li2017learning}
Zhizhong Li and Derek Hoiem.
\newblock Learning without forgetting.
\newblock {\em IEEE transactions on pattern analysis and machine intelligence}, 40(12):2935--2947, 2017.

\bibitem{lin2014microsoft}
Tsung-Yi Lin, Michael Maire, Serge Belongie, James Hays, Pietro Perona, Deva Ramanan, Piotr Doll{\'a}r, and C~Lawrence Zitnick.
\newblock Microsoft coco: Common objects in context.
\newblock In {\em Computer vision--ECCV 2014: 13th European conference, zurich, Switzerland, September 6-12, 2014, proceedings, part v 13}, pages 740--755. Springer, 2014.

\bibitem{liu2022dab}
Shilong Liu, Feng Li, Hao Zhang, Xiao Yang, Xianbiao Qi, Hang Su, Jun Zhu, and Lei Zhang.
\newblock Dab-detr: Dynamic anchor boxes are better queries for detr.
\newblock {\em arXiv preprint arXiv:2201.12329}, 2022.

\bibitem{liu2020multi}
Xialei Liu, Hao Yang, Avinash Ravichandran, Rahul Bhotika, and Stefano Soatto.
\newblock Multi-task incremental learning for object detection.
\newblock {\em arXiv preprint arXiv:2002.05347}, 2020.

\bibitem{liu2023online}
Yaoyao Liu, Yingying Li, Bernt Schiele, and Qianru Sun.
\newblock Online hyperparameter optimization for class-incremental learning.
\newblock In {\em Proceedings of the AAAI Conference on Artificial Intelligence}, volume~37, pages 8906--8913, 2023.

\bibitem{liu2023continual}
Yaoyao Liu, Bernt Schiele, Andrea Vedaldi, and Christian Rupprecht.
\newblock Continual detection transformer for incremental object detection.
\newblock In {\em Proceedings of the IEEE/CVF conference on computer vision and pattern recognition}, pages 23799--23808, 2023.

\bibitem{liu2020mnemonics}
Yaoyao Liu, Yuting Su, An-An Liu, Bernt Schiele, and Qianru Sun.
\newblock Mnemonics training: Multi-class incremental learning without forgetting.
\newblock In {\em Proceedings of the IEEE/CVF conference on Computer Vision and Pattern Recognition}, pages 12245--12254, 2020.

\bibitem{lopez2017gradient}
David Lopez-Paz and Marc'Aurelio Ranzato.
\newblock Gradient episodic memory for continual learning.
\newblock {\em Advances in neural information processing systems}, 30, 2017.

\bibitem{loshchilov2017decoupled}
Ilya Loshchilov and Frank Hutter.
\newblock Decoupled weight decay regularization.
\newblock {\em arXiv preprint arXiv:1711.05101}, 2017.

\bibitem{meng2021conditional}
Depu Meng, Xiaokang Chen, Zejia Fan, Gang Zeng, Houqiang Li, Yuhui Yuan, Lei Sun, and Jingdong Wang.
\newblock Conditional detr for fast training convergence.
\newblock In {\em Proceedings of the IEEE/CVF international conference on computer vision}, pages 3651--3660, 2021.

\bibitem{paik2020overcoming}
Inyoung Paik, Sangjun Oh, Taeyeong Kwak, and Injung Kim.
\newblock Overcoming catastrophic forgetting by neuron-level plasticity control.
\newblock In {\em Proceedings of the AAAI Conference on Artificial Intelligence}, volume~34, pages 5339--5346, 2020.

\bibitem{peng2021sid}
Can Peng, Kun Zhao, Sam Maksoud, Meng Li, and Brian~C Lovell.
\newblock Sid: Incremental learning for anchor-free object detection via selective and inter-related distillation.
\newblock {\em Computer vision and image understanding}, 210:103229, 2021.

\bibitem{pourkeshavarzi2021looking}
Mozhgan PourKeshavarzi, Guoying Zhao, and Mohammad Sabokrou.
\newblock Looking back on learned experiences for class/task incremental learning.
\newblock In {\em International Conference on Learning Representations}, 2021.

\bibitem{prabhu2020gdumb}
Ameya Prabhu, Philip~HS Torr, and Puneet~K Dokania.
\newblock Gdumb: A simple approach that questions our progress in continual learning.
\newblock In {\em Computer Vision--ECCV 2020: 16th European Conference, Glasgow, UK, August 23--28, 2020, Proceedings, Part II 16}, pages 524--540. Springer, 2020.

\bibitem{rebuffi2017icarl}
Sylvestre-Alvise Rebuffi, Alexander Kolesnikov, Georg Sperl, and Christoph~H Lampert.
\newblock icarl: Incremental classifier and representation learning.
\newblock In {\em Proceedings of the IEEE conference on Computer Vision and Pattern Recognition}, pages 2001--2010, 2017.

\bibitem{ren2015faster}
Shaoqing Ren, Kaiming He, Ross Girshick, and Jian Sun.
\newblock Faster r-cnn: Towards real-time object detection with region proposal networks.
\newblock {\em Advances in neural information processing systems}, 28, 2015.

\bibitem{rusu2016progressive}
Andrei~A Rusu, Neil~C Rabinowitz, Guillaume Desjardins, Hubert Soyer, James Kirkpatrick, Koray Kavukcuoglu, Razvan Pascanu, and Raia Hadsell.
\newblock Progressive neural networks.
\newblock {\em arXiv preprint arXiv:1606.04671}, 2016.

\bibitem{shin2017continual}
Hanul Shin, Jung~Kwon Lee, Jaehong Kim, and Jiwon Kim.
\newblock Continual learning with deep generative replay.
\newblock {\em Advances in neural information processing systems}, 30, 2017.

\bibitem{shmelkov2017incremental}
Konstantin Shmelkov, Cordelia Schmid, and Karteek Alahari.
\newblock Incremental learning of object detectors without catastrophic forgetting.
\newblock In {\em Proceedings of the IEEE international conference on computer vision}, pages 3400--3409, 2017.

\bibitem{simon2021learning}
Christian Simon, Piotr Koniusz, and Mehrtash Harandi.
\newblock On learning the geodesic path for incremental learning.
\newblock In {\em Proceedings of the IEEE/CVF conference on Computer Vision and Pattern Recognition}, pages 1591--1600, 2021.

\bibitem{sun2021rethinking}
Zhiqing Sun, Shengcao Cao, Yiming Yang, and Kris~M Kitani.
\newblock Rethinking transformer-based set prediction for object detection.
\newblock In {\em Proceedings of the IEEE/CVF international conference on computer vision}, pages 3611--3620, 2021.

\bibitem{tao2020topology}
Xiaoyu Tao, Xinyuan Chang, Xiaopeng Hong, Xing Wei, and Yihong Gong.
\newblock Topology-preserving class-incremental learning.
\newblock In {\em Computer Vision--ECCV 2020: 16th European Conference, Glasgow, UK, August 23--28, 2020, Proceedings, Part XIX 16}, pages 254--270. Springer, 2020.

\bibitem{wang2022foster}
Fu-Yun Wang, Da-Wei Zhou, Han-Jia Ye, and De-Chuan Zhan.
\newblock Foster: Feature boosting and compression for class-incremental learning.
\newblock In {\em European conference on computer vision}, pages 398--414. Springer, 2022.

\bibitem{wang2025psedet}
Qiuchen Wang, Zehui Chen, Chenhongyi Yang, Jiaming Liu, Zhenyu Li, and Feng Zhao.
\newblock Psedet: Revisiting the power of pseudo label in incremental object detection.
\newblock In {\em The Thirteenth International Conference on Learning Representations}, 2025.

\bibitem{wang2025gcd}
Xu~Wang, Zilei Wang, and Zihan Lin.
\newblock Gcd: Advancing vision-language models for incremental object detection via global alignment and correspondence distillation.
\newblock In {\em Proceedings of the AAAI Conference on Artificial Intelligence}, volume~39, pages 8015--8023, 2025.

\bibitem{yan2021dynamically}
Shipeng Yan, Jiangwei Xie, and Xuming He.
\newblock Der: Dynamically expandable representation for class incremental learning.
\newblock In {\em Proceedings of the IEEE/CVF conference on computer vision and pattern recognition}, pages 3014--3023, 2021.

\bibitem{yoon2019scalable}
Jaehong Yoon, Saehoon Kim, Eunho Yang, and Sung~Ju Hwang.
\newblock Scalable and order-robust continual learning with additive parameter decomposition.
\newblock {\em arXiv preprint arXiv:1902.09432}, 2019.

\bibitem{yoon2017lifelong}
Jaehong Yoon, Eunho Yang, Jeongtae Lee, and Sung~Ju Hwang.
\newblock Lifelong learning with dynamically expandable networks.
\newblock {\em arXiv preprint arXiv:1708.01547}, 2017.

\bibitem{zenke2017continual}
Friedemann Zenke, Ben Poole, and Surya Ganguli.
\newblock Continual learning through synaptic intelligence.
\newblock In {\em International conference on machine learning}, pages 3987--3995. PMLR, 2017.

\bibitem{zhang2022dino}
Hao Zhang, Feng Li, Shilong Liu, Lei Zhang, Hang Su, Jun Zhu, Lionel~M Ni, and Heung-Yeung Shum.
\newblock Dino: Detr with improved denoising anchor boxes for end-to-end object detection.
\newblock {\em arXiv preprint arXiv:2203.03605}, 2022.

\bibitem{zhang2025dynamic}
Jichuan Zhang, Wei Li, Shuang Cheng, Yali Li, and Shengjin Wang.
\newblock Dynamic object queries for transformer-based incremental object detection.
\newblock In {\em ICASSP 2025-2025 IEEE International Conference on Acoustics, Speech and Signal Processing (ICASSP)}, pages 1--5. IEEE, 2025.

\bibitem{zhao2020maintaining}
Bowen Zhao, Xi~Xiao, Guojun Gan, Bin Zhang, and Shu-Tao Xia.
\newblock Maintaining discrimination and fairness in class incremental learning.
\newblock In {\em Proceedings of the IEEE/CVF conference on computer vision and pattern recognition}, pages 13208--13217, 2020.

\bibitem{zhao2024detrs}
Yian Zhao, Wenyu Lv, Shangliang Xu, Jinman Wei, Guanzhong Wang, Qingqing Dang, Yi~Liu, and Jie Chen.
\newblock Detrs beat yolos on real-time object detection.
\newblock In {\em Proceedings of the IEEE/CVF conference on computer vision and pattern recognition}, pages 16965--16974, 2024.

\bibitem{zhu2020deformable}
Xizhou Zhu, Weijie Su, Lewei Lu, Bin Li, Xiaogang Wang, and Jifeng Dai.
\newblock Deformable detr: Deformable transformers for end-to-end object detection.
\newblock {\em arXiv preprint arXiv:2010.04159}, 2020.

\end{thebibliography}
}

\end{document}